\documentclass[10pt,twocolumn,letterpaper]{article}

 \usepackage{cvpr}              
\usepackage{caption}
\usepackage[T1]{fontenc}
\usepackage{textgreek}
\usepackage{threeparttable}

\usepackage[dvipsnames]{xcolor}

\usepackage{indentfirst}
\usepackage{tabularx}
\usepackage{makecell}
\usepackage{stackengine}
\usepackage{booktabs}
\usepackage{pifont}
\newcolumntype{M}[1]{>{\centering\arraybackslash}m{#1}}
\usepackage{threeparttable}

\definecolor{cvprblue}{rgb}{0.21,0.49,0.74}
\usepackage[pagebackref,breaklinks,colorlinks,citecolor=cvprblue]{hyperref}

\def\paperID{2102} 
\def\confName{CVPR}
\def\confYear{2024}

\title{LED: A Large-scale Real-world Paired Dataset for Event Camera Denoising}
\author{Yuxing Duan\textsuperscript{1}, Shihan Peng\textsuperscript{1}, Lin Zhu\textsuperscript{2}, Yi Chang\textsuperscript{1}\footnotemark[1], Wei Zhang\textsuperscript{3}, Sheng Zhong\textsuperscript{1}, Luxin Yan\textsuperscript{1}\\
	\textsuperscript{1}National Key Lab of Multispectral Information Intelligent Processing Technology \\
	\textsuperscript{1}Huazhong University of Science and Technology ,\textsuperscript{2}Beijing Institute of Technology,\textsuperscript{3}Peng Cheng Laboratory\\}
	
\begin{document}
\twocolumn[{%
\renewcommand\twocolumn[1][]{#1}%
\maketitle
\begin{center}
		\centering
		\captionsetup{type=figure}
		\includegraphics[width=\linewidth]{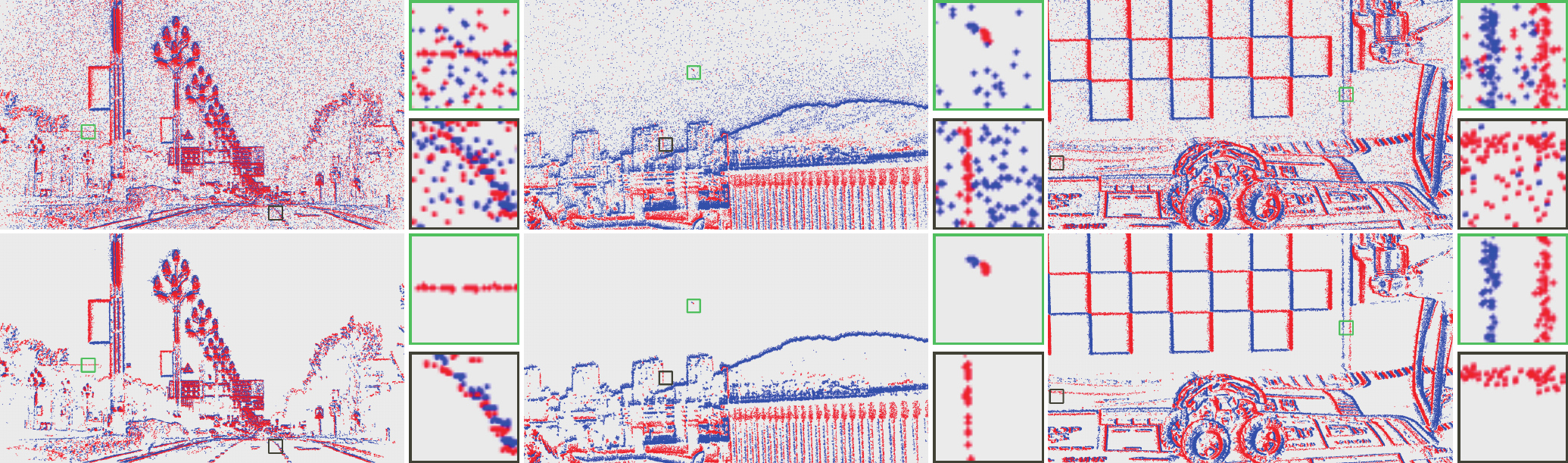}
		\caption{Illustration of the paired raw/denoised event datasets. The proposed real-world event denoising datasets contribute to addressing the inevitable Background Activity (BA) noise encountered by event cameras application under varying camera settings and illumination.}
		\label{fig:fig_1}
  \end{center}%
}]

\setlength{\parindent}{1em}
\begin{abstract}
\vspace{-0.8em}
Event camera has significant advantages in capturing dynamic scene information while being prone to noise interference, particularly in challenging conditions like low threshold and low illumination.
However, most existing research focuses on gentle situations, hindering event camera applications in realistic complex scenarios.
To tackle this limitation and advance the field, we construct a new paired real-world event denoising dataset (LED), including 3K sequences with 18K seconds of high-resolution (1200*680) event streams and showing three notable distinctions compared to others: diverse noise levels and scenes, larger-scale with high-resolution, and high-quality GT.
Specifically, it contains stepped parameters and varying illumination with diverse scenarios.
Moreover, based on the property of noise events inconsistency and signal events consistency, we propose a novel effective denoising framework(DED) using homogeneous dual events to generate the GT with better separating noise from the raw.
Furthermore, we design a bio-inspired baseline leveraging Leaky-Integrate-and-Fire (LIF) neurons with dynamic thresholds to realize accurate denoising.
The experimental results demonstrate that the remarkable performance of the proposed approach on different datasets.The dataset and code are at \url{https://github.com/Yee-Sing/led}.
\end{abstract}
\vspace{-1em}   
\crefname{table}{Table}{Tables} 
\begin{table*}[t]
	\setlength{\abovecaptionskip}{5pt}
	\setlength{\belowcaptionskip}{-0.4cm}
	\centering
	\renewcommand{\arraystretch}{0.80} 
	\fontsize{8}{12}\selectfont 
	\begin{tabular}{lM{0.07\linewidth}M{0.07\linewidth}M{0.12\linewidth}M{0.09\linewidth}M{0.18\linewidth}M{0.08\linewidth}M{0.04\linewidth}}
		\toprule
		Datasets & Sequences & Capture/s & Source & Noise level & Illumination & Resolution & Paired \\
		\midrule
		RGB-DAVIS \cite{wang2020joint} & 20 & 122 & Camera & \textcolor{green}{\ding{55}} &  Daytime & 180*190 &\textcolor{green}{\ding{55}} \\
		EventNFS \cite{duan2021eventzoom}  & 100 & 4238 & Screen Recording & \textcolor{green}{\ding{55}} & Daytime & 224*124 &\textcolor{green}{\ding{55}} \\
		E-MLB \cite{ding2023mlb} & 1200 & 7300 & Camera & 4 & Daytime to Nighttime & 346*260 &\textcolor{green}{\ding{55}} \\
		\midrule
		DVSNOISE20 \cite{baldwin2020event} & 48 & 807 & Camera & \textcolor{green}{\ding{55}} & Daytime & 346*260 & \textcolor{red}{\ding{51}} \\
		ED-KoGTL \cite{alkendi2022} & 4 & 15 & Camera & \textcolor{green}{\ding{55}} & Daytime to Nighttime & 346*260 &\textcolor{red}{\ding{51}}\\
		DVSCLEAN \cite{fang2022aednet} & 144 & 55 & Synthesis & 2 & Daytime & 1280*720 &\textcolor{red}{\ding{51}}\\
		\textbf{LED} & 3000 & 18000 & Camera & 4 & Daytime to Nighttime & 1200*680 &\textcolor{red}{\ding{51}} \\
		\addlinespace[-0.5ex] 
		\bottomrule
	\end{tabular}
	\caption{Summary of existing event denoising datasets.}
	\label{tab:tab_1}
\end{table*}
\section{Introduction}
\label{sec:intro}
Event cameras possess unique imaging advantages and are demonstrating tremendous potential for various applications \citep{falanga2020dynamic,perot2020learning,mangalwedhekar2023achieving,cabriel2023event}.
Unlike traditional frame cameras using integration sampling \cite{wei2020physics}, event camera is differential sampling \cite{lichtsteiner2008}, evading the restriction of exposure period.
However, due to this sampling mechanism, random fluctuations in analog signals can easily form noise events.
In contrast to the dominance of signal components in images, the noise and signal in event cameras are encoded with same magnitude, easily disturbing inherent structural features, such as lane lines shown in \cref{fig:fig_1}.
BA noise detrimentally impact on subsequent tasks like reconstruction \citep{rebecq2019high,pan2020high}, motion estimation \citep{stoffregen2019event,stoffregen2019reward}and event-based object detection \cite{chen2020end}.
Therefore, event denoising becomes a fundamental issue, notably in the booming event-based vision community.

Paired data is a key issue for event denoising in the deep-learning era, thus an intuitive idea is to construct paired data through simulators  \citep{hu2021v2e,gehrig2020video,rebecq2018esim}.
However, the image-based simulation cannot be equivalent to reality because of the domain gap.
To obtain the `clean part' from the real-world event, multi-modal data was incorporated to help paired event generation.
APS images gradient combined with IMU are used to indicate the event probability within frame period \citep{baldwin2020event}.
By jointly estimating the motion parameters with APS images and event streams \citep{wang2020joint,duan2021guided}, the temporal mismatch can be alleviated.
Constructing a fixed-track setup \cite{alkendi2022}, the events under various illuminations can be referred to APS images in good lighting.
Essentially, these methods align with events via transformed multi-mode features.
Unfortunately, this heterogeneous aided information cannot fully match with the event, and their effectiveness is unavailable in situations such as low light and motion blur.
Hence, few datasets have considered practical scenes where BA noise is easily deteriorated by camera setting and illumination, appearing as complex and nonuniform distributions.

To address this issue, we construct a large-scale, high-quality paired dataset LED for event denoising.
LED has three distinct advantages. Firstly, it contains diverse noise levels, covering complex situations under various camera settings and illumination.
Secondly, LED includes such as outdoor and indoor scenarios, with abundant objects.
Last but not least, LED is collected with the current largest-scale and high-resolution.
Moreover, inspired by the multi-sampling images denoising methods \citep{mildenhall2018burst,hasinoff2016burst,liu2014fast,liba2019handheld}, we are the first to explore event multi-sampling for denoising.
Based on the homogeneous data, we proposed a dual-events denoising framework that can generate higher-quality GT by more accurately separating BA noise from the raw.    

The contributions of this paper can be summarized as:

(1) We provide a large-scale real-world paired dataset for event denoising. So far, \textbf{LED} is the largest paired real-world event denoising dataset (3000 sequences) and high-resolution (1200*680) with diverse noise levels and scenes. 

(2) We propose a novel dual events denoising framework \textbf{DED} based on noise inconsistency. We provide detailed analysis to show that DED can better separate mixed signals and noisy event streams across various camera threshold parameters, light conditions and motion patterns.

(3) We introduce a novel baseline \textbf{DTSNN} for event denoising based on the spiking neural network, which utilizes the learnable dynamical spike threshold of the LIF neuron to accurately denoising. Extensive experiments on different real datasets verify the superiority of the proposed method. 

\section{Related work}
\label{sec:Related work}

\begin{figure*}[t]
	\setlength{\abovecaptionskip}{5pt}
	\setlength{\belowcaptionskip}{-0.25cm}
	\centering
	\includegraphics[width=\linewidth]{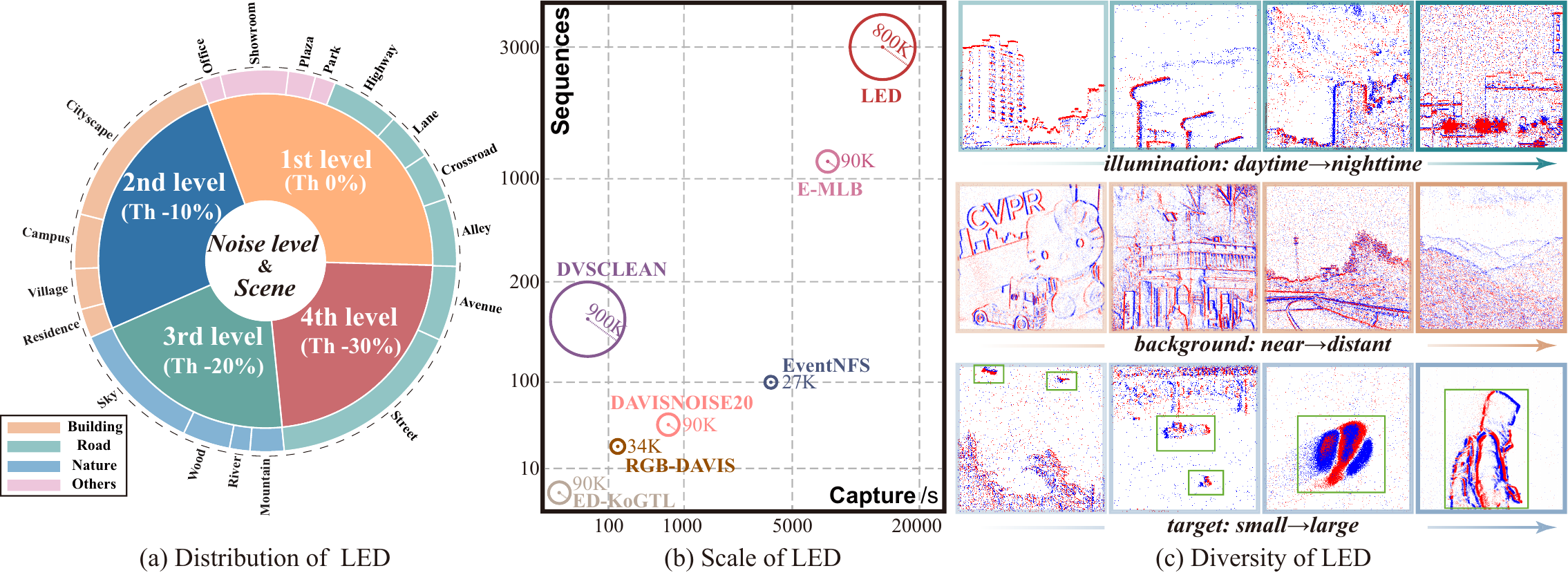}
	\caption{ Illustration of the proposed dataset LED. (a) Distribution of noise level and scene of the proposed dataset. (b) Our proposed LED outperforms others in terms of sequences,  capture, and resolution (Circles with numbers to indicate). (c) LED collects diverse event streams across various conditions of illumination, depth of field, and target scale.}
	\label{fig:fig_4}
\end{figure*}

\begin{figure}
	\setlength{\abovecaptionskip}{5pt}
	\setlength{\belowcaptionskip}{-0.35cm}
	\centering
	\includegraphics[width=0.75\linewidth,height=5.2cm]{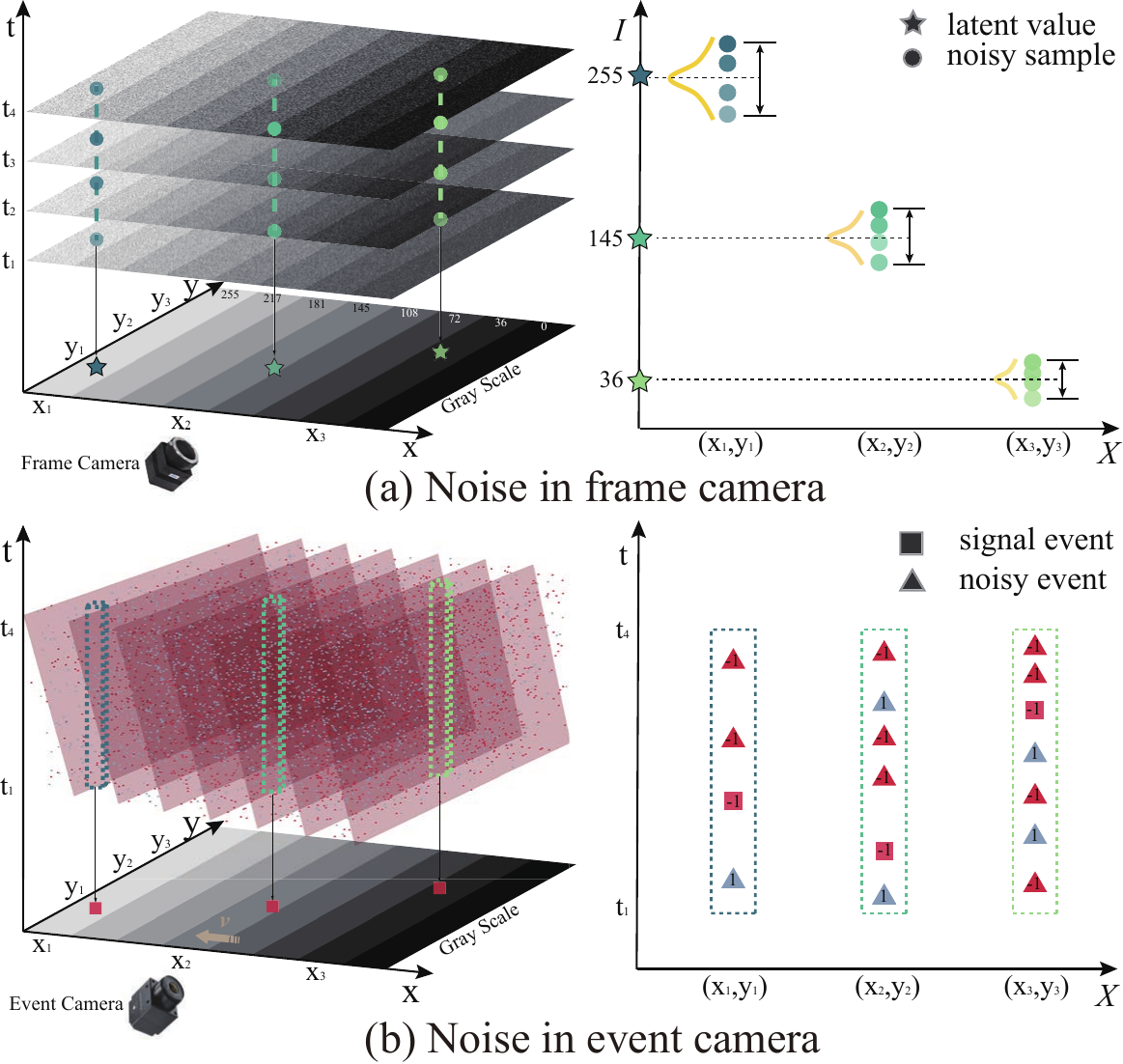}
	
	\caption{The substantial noise distinctions between frame/event camera. (a) For a stationary grayscale chart, the frame camera sequentially acquires noisy samples, fluctuating around a latent value within a specific distribution. (b) In a horizontal moving case, the event camera outputs binary signal serially, comprising signal events from motion gradient edges and spurious BA noise.}
	\label{fig:fig_2}
\end{figure}

\begin{figure*}
	\setlength{\abovecaptionskip}{5pt}
	\setlength{\belowcaptionskip}{-5pt}
	\centering
	\includegraphics[width=\linewidth]{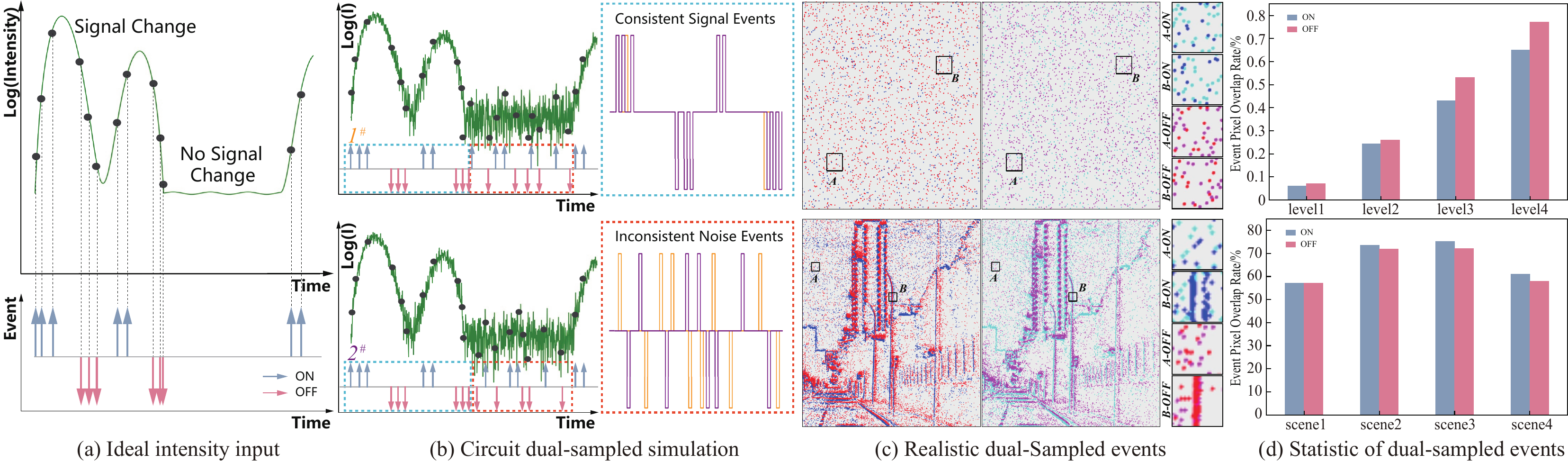}
	\caption{Illustration of event camera dual-sampling analysis:
		(a) Given an ideal intensity input, which includes varying and steady stages, generates a series of signal events.
		(b) The twice samplings of the same input in the circuit model, both generate additional BA noise in previously steady stages, resulting in consistency discrepancies between noise events and signal events.
		(c) The visualization results of the actual dual-sampled events cumulative frame demonstrate the misalignment of inconsistent dual-sampled noise events and the alignment of coexisting signal events (two color groups indicating the respective polarities of the dual events).
		(d) The statistical results of the two tests also prove a low overlap rate between dual-sampled event pixels and a much higher one where signal events are present.}
	\label{fig:fig_3}
\end{figure*}
\noindent\textbf{Event Denoising Datasets}. 
Currently, the datasets exclusively designed for event denoising with less attention compared to the task-specific ones in the event-based community.
In \cref{tab:tab_1}, we provide an all-sided summary of existing datasets, highlighting the scarcity of real paired datasets.

To address the lack of explicit labels, some paired datasets can be conveniently generated via simulators such as DVSCLEAN \cite{fang2022aednet} and DND21 \cite{guo2022low}.
However, due to the gap between simulation and real, these simulators fail to fully reflect the actual situation.
Other works focus on real paired event denoising data.
DVSNOISE20 \cite{baldwin2020event} captures data in 16 pure background scenes.
RGB-DAVIS \cite{wang2020joint} provides 20 indoor/outdoor sequences under good lighting conditions.
ED-KoGTL \cite{alkendi2022} constructs indoor data with four illumination levels under constant trajectory.
These paired datasets have achieved favorable results, primarily attributable to the gradient cues extracted from APS images.
However, the heterogeneity between images and events limits the availability of diverse lighting conditions, motion patterns, and scenarios for constructing paired data.

\noindent\textbf{Event-based Denoising.} Research on event denoising can be categorized into two approaches: filtering-based methods and deep learning methods. The former relies on manual priors to design discriminative models for noise removal, such as utilizing density distinction \cite{liu2015design,guo2022low,feng2020event,zhang2023neuromorphic}or motion continuity distinction \cite{benosman2013event,wang2019ev,wu2020probabilistic} and motion compensation \cite{chen2022progressivemotionseg,wang2020joint} with CM framework \cite{gallego2018unifying} in the spatiotemporal domain. However, the validity of this prior knowledge is sensitive to different signal/noise event distributions, limiting their denoising accuracy. For learning-based methods, such as the CNN-based EDNCNN \cite{baldwin2020event}, the PointNet-based AEDNet \cite{fang2022aednet} and \cite{alkendi2022}  GNN-transformer were proposed successively. Additionally, the event denoising and super-resolution tasks are combined in \cite{duan2021eventzoom}.

\noindent\textbf{Multi-Sampling Image Denoising}. The multi-sampling principle for image denoising has been extensively studied.
Techniques like long exposure time or burst captures benefit the direct acquisition of clear data with increased SNR in static scenes.
Further, the multi-sampling was explored to the paradigm of indirectly recovering noise-free data.
Not only using paired noisy observations to learn the mapping from different noisy samples to clear counterpart \cite{lehtinen2018noise2noise,maleky2022noise2noiseflow}, but also employing spatial sub-sampling on a single noisy image to learn the mapping from adjacent noisy samples to GT \cite{krull2019noise2void,huang2021neighbor2neighbor}. 
Evidently, these denoising approaches imply that signal resampling could facilitate suppressing random circuit noise and approximating the expectation value.
\section{The High-quality Paired LED Dataset} 
\label{sec:Method}

\subsection{Dataset Statistics and Features}
The primary factors affecting BA noise in event cameras can be categorized into intrinsic camera settings and extrinsic lighting conditions.
By setting different camera thresholds, we can selectively record events with varying noise levels.
Generally, a lower threshold means higher sensitivity but also increases the likelihood of BA noise.
In contrast to fixed parameters in other datasets, we use four level thresholds to collect events, starting from the default value of 0\% and gradually decreasing to -10\%, -20\%, and -30\%, as shown in \cref{fig:fig_4} (a).
To cope with the challenges posed by complex lighting conditions, data collection is from daytime to nighttime, encompassing a transition from high to low illumination and a mixture of natural light with artificial light sources.
Thus, the impact of external conditions on the  event stream distribution can also be controlled.

Due to the difficulty of constructing real paired datasets, the development of existing datasets remains limited, as shown in \cref{tab:tab_1}. 
In this work, we used a vehicle-mounted or pan-tilt equipped with event cameras featuring a 16mm lens.
Totally, we collected approximately 5 hours of event stream, including 3K sequences.
The composition of noise levels and scenes is depicted in \cref{fig:fig_4} (a), with roughly equal amounts collected during daytime and nighttime.
Nearly 20 typical scenes were captured, including urban buildings, traffic roads, natural scenery, indoor exhibits, etc.

Our dataset goes beyond just the number of sequences and data volume, primarily aiming to consider the intricate influences of relevant factors on noise.
Thus, we endow LED with diversity to reflect the impact of objects and environments on the event, which helps cover the cases of densely and non-uniformly distributed noisy situations in reality both spatially and temporally.
In \cref{fig:fig_4} (c), we exhibit that LED diversity is not only embodied in the richness of noise and illumination levels but also features such as depth of field and scene semantics.
Meanwhile, high-resolution imaging aids the event cameras in capturing spatial information more precisely.
Unfortunately, most existing ones with low resolution may hinder advanced visual tasks.
Therefore, we collected data with 1200*680 resolution(after cropping), achieving high spatiotemporal resolution.
Additionally, LED includes abundant objects of various scales, including aerial targets, pedestrians, and vehicles, making it well-suited for downstream tasks.
\begin{figure}[t]
	\setlength{\abovecaptionskip}{5pt}
	\setlength{\belowcaptionskip}{-5pt}
	\centering
	\includegraphics[width=1.07\linewidth,height=3.6cm]{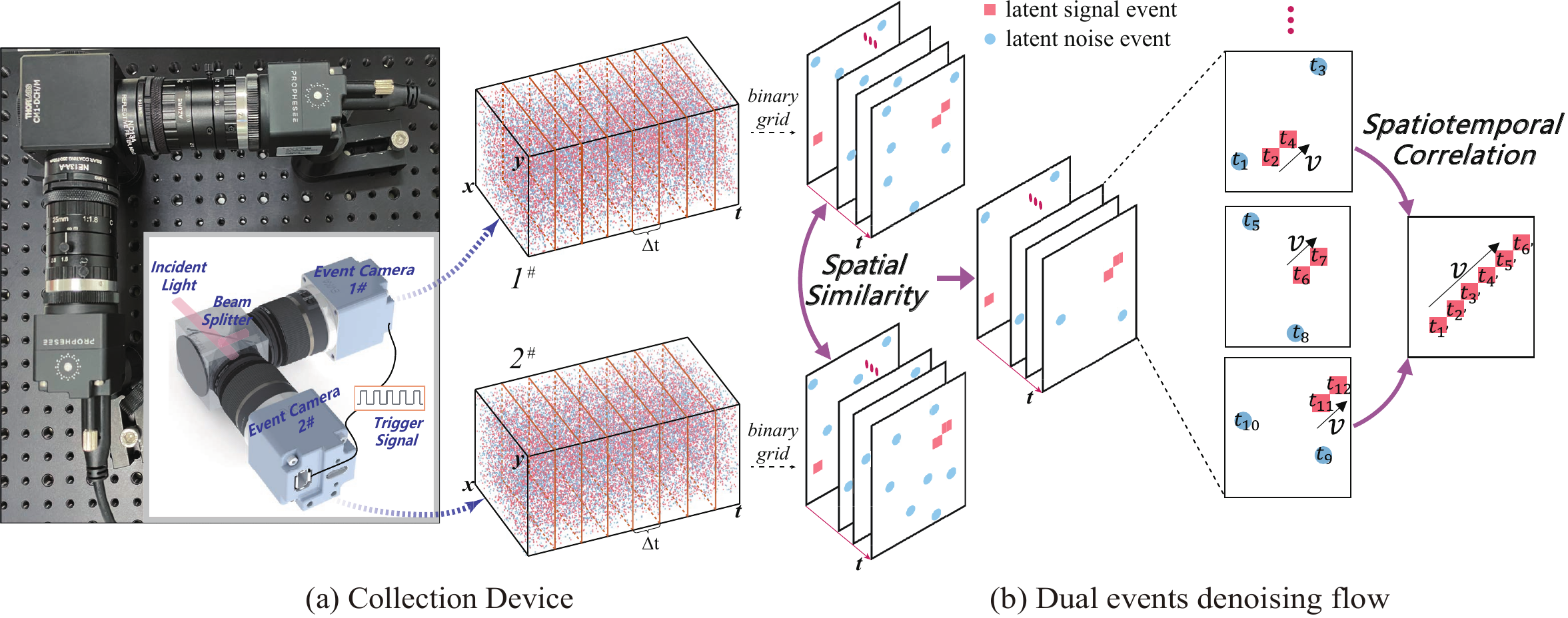}
	\caption{Overview of the dual events denoising framework. (a) Our collection device consists of two identical EVK4s forming a co-axial system with a 1:1 beamsplitter. (b) We first perform spatial similarity processing to retain the consistent parts, followed by sequentially spatiotemporal correlation constraints to remove the residual small amount of noise event from the previous step.}
	\label{fig:fig_5}
\end{figure}
\subsection{Dual Events Denoising Framework}
Given a noisy event stream, the key lies in properly obtaining paired GT.
In this section, we explore utilizing multi-sampling to achieve event stream denoising.

\noindent\textbf{Background}. Multi-sampling denoising methods are commonly used in image denoising.
For frame images, averaging the results of temporal multi-samples can approximate the true value for each pixel owing to multi-samples smoothing effect on random noise components, as shown in \cref{fig:fig_2} (a).
For event stream, due to their dynamic sampling and the binary values, temporal multi-samples may contain both signal events and BA noise, as illustrated in \cref{fig:fig_2} (b), directly taking the average of multi-samples cannot yield true signal value.
Although the noise forms differ between the two modalities, they ultimately originate from the inherent noise in the circuitry analog signal, while resampling fundamentally helps suppress the random noise source.
\begin{figure*}[htbp]
	\setlength{\abovecaptionskip}{5pt}
	\setlength{\belowcaptionskip}{-0.4cm}
	\centering
	\includegraphics[width=\linewidth]{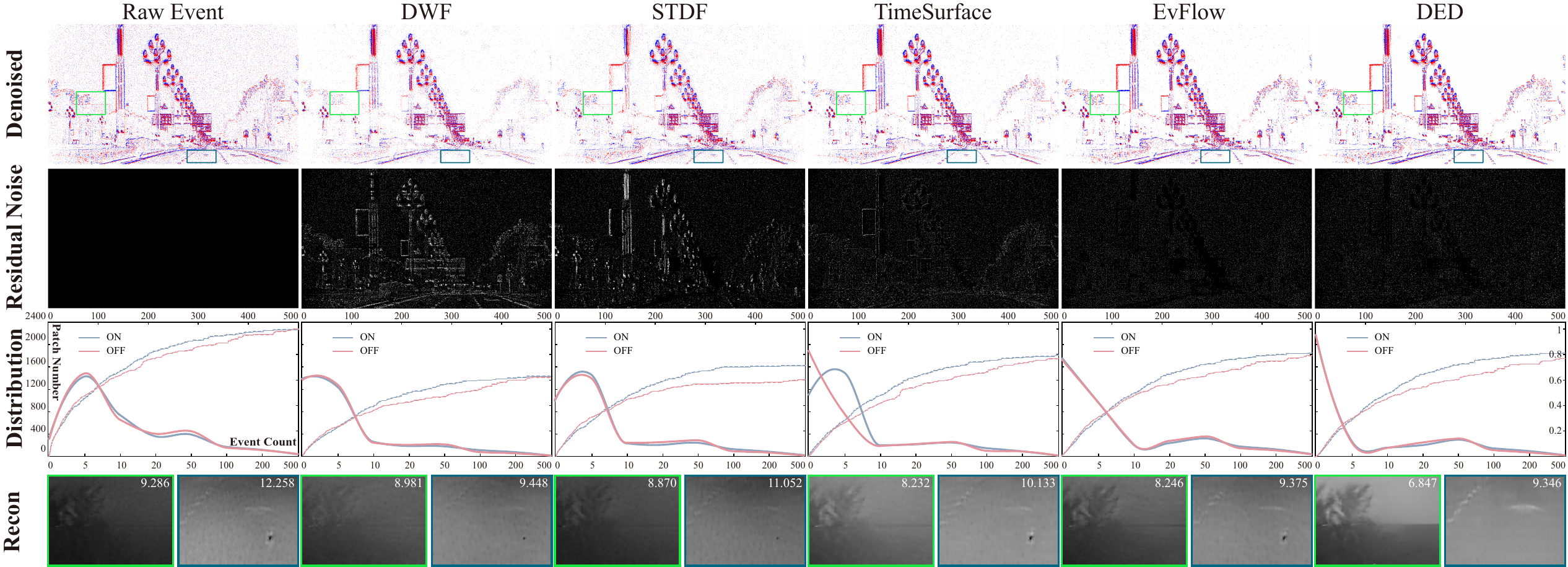}
	\caption{Analysis of different event denoising results on our dataset. From left to right, the first column is the raw events, and the remaining five columns represent different methods, namely DWF, STDF, TimeSurface, EvFlow, and the proposed DED. From top to bottom, the first row shows the denoised results, the second row is the residual noise, the third row denotes the statistical distribution of the denoised results and the last row represents the intensity image reconstruction corresponding to the zoomed region denoised events.}
	\label{fig:fig_6}
\end{figure*}

\noindent\textbf{Analysis}. This concept inspired us to utilize multi-sampling for event denoising, particularly when considering the binary state of events (presence or absence) where just two samplings are required to potentially identify noise.
Thus, we initially investigated the noise inconsistency at the pixel level.
Based on the DVS working principle \cite{lichtsteiner2008}, we build a circuit model in the Cadence \cite{cadence} platform to simulate the dual-sampling with twice identical intensity input. 
The noisy analog signal converted from input generates BA noise at the originally steady stage compared to the ideal case in \cref{fig:fig_3} (a).
Clearly, the two sets of noise exhibit inconsistency, while the signals demonstrate high consistency, causing distinct alignment differences between the two cases as shown in the close-up of \cref{fig:fig_3} (b).

To further explore the inter-class differences between noise and signal events in real-world dual-sampled events stream, we established a collection device as shown in \cref{fig:fig_5} (a), to construct genuinely spatiotemporal dual-sampling.
Two experimental conditions to validate the aforesaid circuit simulation phenomenon are a stationary camera and a moving one respectively.
In the former, pure BA noise events was obtained, while the latter captured mixed events containing both noise and signal.
In static tests, although at the highest noise level (-30\% threshold), both two sets of events exhibit significant spatiotemporal disparity.
As observed in the right zoomed-in patches at the upper row in \cref{fig:fig_3} (c), the global spatial misalignment of dual BA noises events was evident.
The overlap rate of the pixels triggering event is also to be less than 1\%, as shown in the upper-row histogram of \cref{fig:fig_3} (d).
Not unexpectedly, in dynamic tests, regions with BA noise events remain inconsistency, while the coexistent signal events showing spatiotemporal alignment as depicted in lower-row counterpart, leading to a dramatic increase in overlap rate compared to the pure BA noise cases, as shown in the bottom histogram of \cref{fig:fig_3} (d).

\noindent\textbf{Formulation}. These results support the insight that the consistency discrepancies between dual-sampled events could naturally help distinguish noise.
Therefore, we develop a denoising framework with the dual events stream, called DED. In DED framework, given the spatiotemporal synchronized dual event streams $Y_{1}$ and $Y_{2}$, they are composed of common latent signal $X$ and respective noise streams $N_{1}$ and $N_{2}$. The noise model can be defined as follows:
\begin{equation}
	\setlength{\abovedisplayskip}{2pt}
	\setlength{\belowdisplayskip}{2pt}
	Y_{1} = X + N_{1} \enspace,\enspace Y_{2} = X + N_{2}. 
	\label{noise_model}
\end{equation}
Because of the consistency of signal events and the inconsistency of noise events in $Y_{1}$ and $Y_{2}$, after binary grid operation of event stream according to certain temporal window $\Delta t$, we can perform spatial similarity to process them:
\begin{equation}
	\setlength{\abovedisplayskip}{2pt}
	\setlength{\belowdisplayskip}{2pt}
	X^* = \left\|  \left\| X + N_1 \right\| \circ \left\| X + N_2 \right\| \right\|_1,
	\label{eq:ss}
\end{equation}
where $\circ$ denotes the Hadamard product used to obtain the similar part $X^*$ of two groups of raw event binary frames.
Due to the probability of a few noises occurring simultaneously in dual-sampling, $X^*$ may still contain some much-isolated noise compared to the raw.
As shown in \cref{fig:fig_5} (b), for finer denoising, we further utilize the spatiotemporal correlation of the signal event stream to remove the residual noise from the previous step.
Specifically, we accumulate the events from the relevant spatiotemporal range in corresponding $X^*$ within several consecutive $\Delta t$ to form a spatiotemporal relationship set. The presence of these few isolated noise events, which are randomly triggered in the spatiotemporal domain, disrupts the inherent spatiotemporal correlation of signal reflecting a certain motion model. Therefore, to leverage this spatiotemporal continuity, we transform it into the following minimization problem:
\begin{equation}
	\setlength{\abovedisplayskip}{2pt}
	\setlength{\belowdisplayskip}{2pt}
	\begin{aligned}
		\min & \frac{1}{{N-1}} \sum_{i=1}^{N} \left| e_{\boldsymbol{x}}^{t_{i+1}} - e_{\boldsymbol{x}}^{t_{i}} \right|, \substack{t_i \in  n\Delta t, \\ {\boldsymbol{x}} \in \Omega, \Omega \in X^*},
	\end{aligned}
	\label{eq:sc}
\end{equation}
where $e_{\boldsymbol{x}}^{t_{i}}$ denotes the $i_{th}$ occurring event according to the sorted timestamp, $\boldsymbol{x}$ means the event spatial coordinates, $ n $ is the number of selected consecutive time windows, $\Omega$ represents the candidate spatial neighborhood, and $N$ is total events number in $\Omega$. Additionally, the aforesaid processing is all performed on the two channels of event polarity.

\subsection{Evaluation and Discussion}
The quality of GT is critical for a real paired event dataset.
However, it is challenging to objectively evaluate the denoising results since directly obtaining clean data is unpractical.
To fairly assess the denoising effects of different methods, we conducted a comprehensive evaluation.

\cref{fig:fig_6} presents comparative results of representative event denoising methods could be used for GT generation: spatiotemporal density filter-based (DWF\citep{guo2022low}, STDF\citep{feng2020event}), smoothness optimization-based (TimeSurface\citep{Lagorce2017}), and motion estimation-based (EvFlow\citep{wang2019ev}).
The first and second rows display the denoised and residual noise visualization, respectively.
It can be observed that DED effectively eliminates almost all the scattered BA noise with good preservation of the inherent structured feature, while the others exhibit residual noise and varying degrees of signal event damage.
Statistical analysis on the global distribution of the denoised event is shown in the third row.
Compared to others, DED demonstrates a higher number of blank patches while maintaining a high event preservation rate, which better reflects the overall sparsity and local concentration of the ideal event stream.
Moreover, better denoising often leads to higher-quality reconstruction.
In the last row, the intensity images reconstructed by E2VID \citep{rebecq2019high} indicate that DED accurately restores nighttime scene information, such as leaves, building contours, and road signs with lower NIQE \citep{mittal2012making}, which further indirectly validates the accuracy of denoised results and also confirms the superior effect of the processed data on reconstruction.
\vspace{-1em}
\label{sec:Evaluation and Discussion}
\begin{figure}[t]
	\setlength{\abovecaptionskip}{5pt}
	\setlength{\belowcaptionskip}{-0.4cm}
	\centering
	\includegraphics[width=\linewidth,height=6 cm]{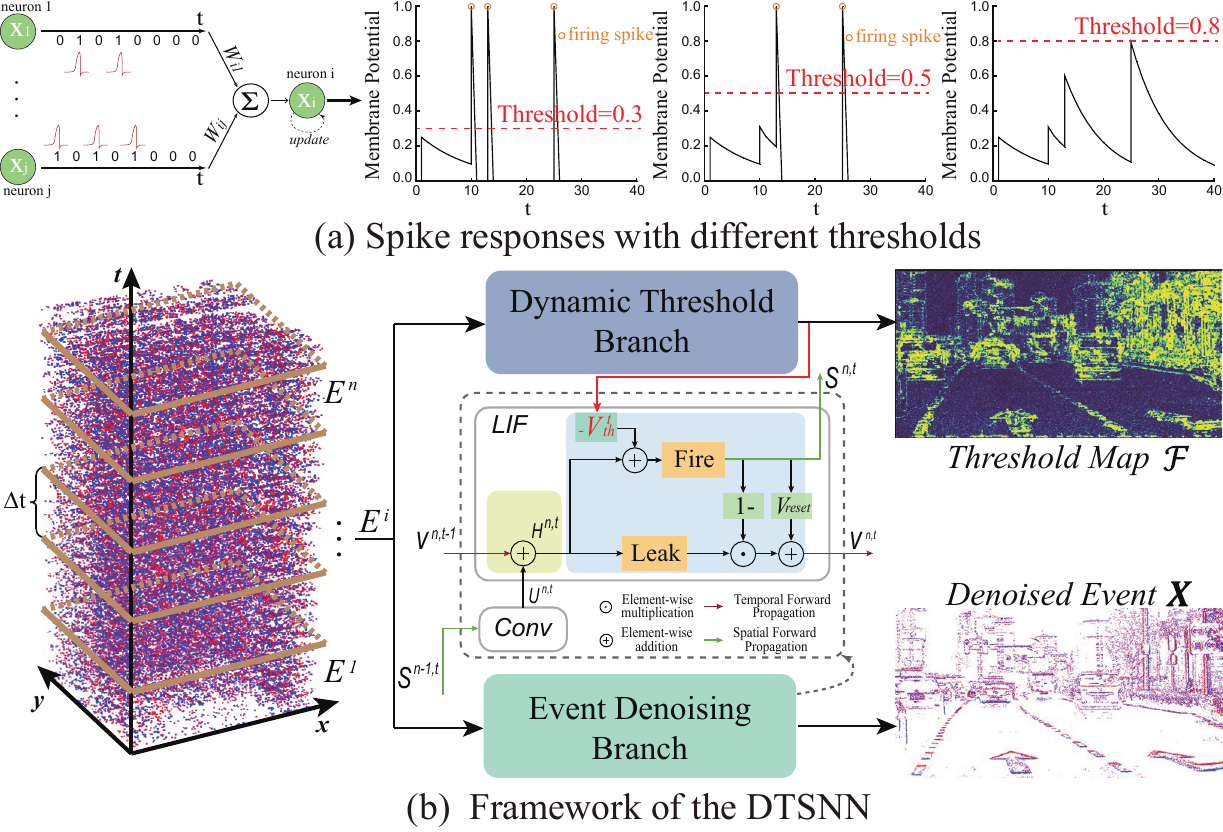}
	\caption{An overview of the DTSNN model. (a) The dynamic threshold mechanism of LIF neuron. (b) The model with two branches consisting of DTB and EDB. Please refer to our supplementary material for more  details about network architecture.}
	\label{fig:fig_7}
\end{figure}
\vspace{-0.3em}
\section{DTSNN for Event Denoising}
\label{sec:DTSNN}
Compared to artificial neural networks(ANN), spiking neural network (SNN) have lower precision but are inherently suited for processing event-driven data due to their  and information transfer mode temporal dynamics, gradually gaining prominence in event-based tasks \cite {yao2021temporal,zhang2022spiking,zhu2022event,zhang2021event}.

\noindent\textbf{Dynamic Threshold Mechanism}. Adjustable spiking neuron offers enhanced biological plausibility \cite{fang2021incorporating,ding2022biologically,wang2022ltmd}.
Indeed, the neuron firing threshold, similar to the event camera threshold parameter, functions as controlling the output.
Intuitively, the dynamic threshold (DT) mechanism can adapt to more complicated situations since a high threshold could suppress noisy input when noise dominates, while a low one helps to sensitively preserve the expected signals.
Inspired by this, we transformed the fixed threshold into dynamic ones based on the LIF neuron model \cite{gerstner2014neuronal} to mimic this anisotropic biomechanism. The tendency of decreasing spike quantity of the postsynaptic neurons responses with an increasing threshold as illustrated in \cref{fig:fig_7}(a).

\noindent\textbf{DTSNN Model}.
Therefore, we propose a fully SNN (all synaptic operations are SNN-based) incorporating learnable DT for event denoising, namely DTSNN.
As illustrated in \cref{fig:fig_7} (b), DTSNN consists of a dynamic threshold branch (DTB) and event denoising branch (EDB).
The former dynamically generates a threshold map based on the successive event input, indicating the approximate spatial regions of signal or noise.
This map is then passed to the latter, controlling the spiking process of LIF neurons in the last layer, which is depicted in \cref{fig:fig_7} (b) and can be formulated as:
\begin{equation}
	\left\{
	\begin{aligned}
			H^{n,t}  &=  V^{n,t-1} + U^{n,t} \\
			S^{n,t}  &=  Hea(H^{n,t}-\boldsymbol{\mathcal{F}}(V_{\text{th}}^t)) \\
			V^{n,t}  &= V_{\text{reset}}S^{n,t}+\frac{1}{\tau}H^{n,t} \odot (1-S^{n,t})
		\end{aligned}
	\right.
	\label{equ:DT}
\end{equation}
where $n$ and $t$ denote the layer number and time step respectively. $H^{n,t}$ is the membrane potential which is produced by coupling the spatial feature $U^{n,t}$ and the temporal input $V^{n,t-1}$. 
The DT map $\boldsymbol{\mathcal{F}}(V_{\text{th}}^t)$ determines whether the output spiking
matrix $S^{n,t}$ should be fired or stay as zero, formulating the final denoised event matrix $\boldsymbol{X}$.
$Hea(x) = 1$ represents the Heaviside step function when $x$ $\geq 0$, otherwise $Hea(x) = 0$, and $\odot$ means
a element-wise multiplication.

Notably, both two branches receive the same consecutive data as input, which is a binary event frame within a certain time window of \textit{\textDelta t}.
The denoising branch is supervised by the GT of signal events within \textit{\textDelta t}, while the threshold map label of the dynamic threshold branch comes from the signal events over a longer temporal period.
\vspace{0.5em}
\begin{figure*}[t]
	\setlength{\abovecaptionskip}{5pt}
	\setlength{\belowcaptionskip}{-5pt}
	\centering
	\includegraphics[width=\linewidth]{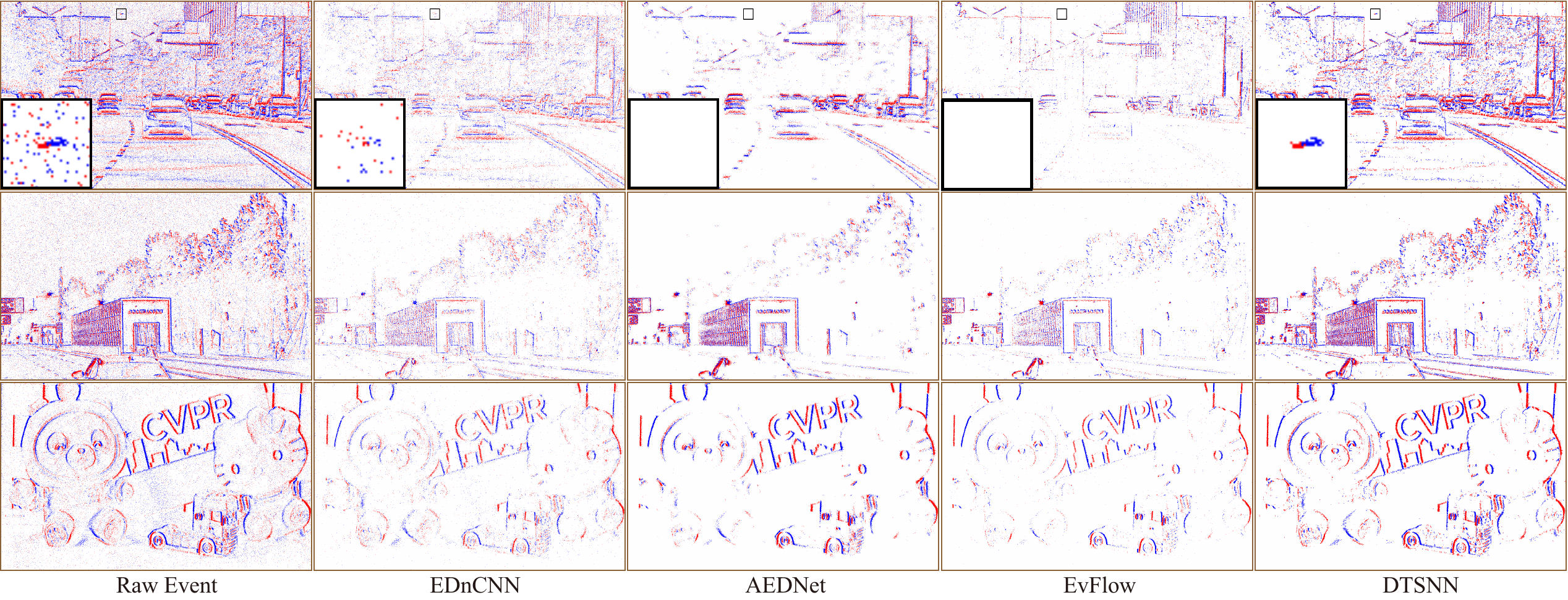}
	\caption{Visual comparisons on LED. Comparing with state-of-the-arts, DTSNN achieves excellent denoising results and it is capable of removing noise events scattered around in sky, ground, and structural contours, simultaneously better  preserve scene details.}
	\label{fig:fig_8}
\end{figure*}

\section{Experiments}
\subsection{Experiments Setup}
\noindent\textbf{Implementation Details}. We use a combined loss consisting of the L1-norm and BCE to train the proposed network.
Our models are implemented with the open-source framework SpikingJelly \cite {fang2023spikingjelly}, using NVIDIA A100 GPU.
The Adam optimizer is employed with a batch size of 8 and a learning rate of 0.002.
A fixed threshold of 0.5 is chosen for the rest of the neurons except at the last membrane potential layer.
Besides, the reset value ${V_{\text{reset}}}$ and the membrane time constant $\tau$ of all LIF neurons are set to 0 and 2, respectively.

\noindent\textbf{Datesets}. We conduct quantitative experiments on existing paired datasets. Due to the enormous scale of billions of events in LED, training them all would be extremely time-consuming and resource-intensive.
Therefore, 600 sequences were randomly selected for training with each consisting of consecutive 10 segments, and 60 sequences were randomly chosen for testing.
To further evaluate the performance of the discrepancy of different datasets, we qualitatively test on typical public datasets DSEC \cite{Gehrig21ral} and E-MLB \cite{ding2023mlb}. For event-based denoising methods, besides the methods in Sec.\ref{sec:Evaluation and Discussion} we select the representative supervised denoising methods including EDnCNN \cite{baldwin2020event} and AEDNet\cite{fang2022aednet}.
 
\noindent\textbf{Evaluation Methods}. We use an index of event denoising accuracy $DA =\frac{1}{2}(\frac{TP}{GP}+\frac{TN}{GN})$ to measures the denoising performance on LED, where TP, TN, GP and GN are the number of true signal, true noise, total signal, and total noise respectively, forming two parts: signal retain (SR) and noise removal (NR). The metrics on other paired datasets are adopted from their proposed ones. Moreover, visualization results was qualitatively evaluated on other datasets. 

\begin{table}[t]
	\setlength{\abovecaptionskip}{5pt}
	\setlength{\belowcaptionskip}{-0.3cm}
	\centering
	\renewcommand{\arraystretch}{0.80}
	\fontsize{8}{12}\selectfont 
	\begin{tabularx}{\linewidth}{@{}lM{0.04\linewidth}M{0.04\linewidth}M{0.04\linewidth}M{0.12\linewidth}M{0.125\linewidth}M{0.135\linewidth}@{}}
	\toprule
		Datasets & \multicolumn{3}{c}{LED} & DVS CLEAN & DVS NOISE20 & Average\\[0.15cm] 
		\cmidrule{1-1} \cmidrule{2-4} \cmidrule{5-5} \cmidrule{6-6} \cmidrule{7-7}		
		Metrics & SR$\uparrow$ & NR$\uparrow$ & DA$\uparrow$ & SNR$\uparrow$ & RPMD$\downarrow$ &Runtime$\downarrow$\\
		\midrule
		Knoise\cite{Knoise2018} & 16.4 & \underline{98.9} & 57.6 & 25.21 & 17.64 & \textbf{1.33} $ms$ \\
		DWF\citep{guo2022low} & 42.6 & 83.9 & 63.3 & 26.96 & 31.39 & 3.67 $ms$ \\
		STDF\citep{feng2020event} & 37.4 & \textbf{99.1} & 68.3 & 19.30 & 30.23 & \underline{2.04} $ms$ \\
		TS\citep{Lagorce2017}  & 30.6 & 98.3 & 64.5 & 13.98 & \textbf{15.25} & 5.76 $ms$ \\
		EvFlow\citep{wang2019ev} & 52.3 & 96.4 & 74.4 & 23.74 & 22.50 & 150.5 $ms$ \\
		\midrule
		EDnCNN\cite{baldwin2020event} & 80.0 & 82.0 & 81.0 & 20.29 & 22.25 & 251.4 $ms$ \\
		AEDNet\cite{fang2022aednet} & \underline{81.2} & 83.6 & \underline{82.4} & \underline{25.58} & 18.51 & 708.1 $ms$ \\
		\textbf{DTSNN} & \textbf{86.0} & 86.5 & \textbf{86.2} & \textbf{29.26} & \underline{16.13} &8.20 $ms$ \\
		\addlinespace[-0.5ex] 
		\bottomrule
	\end{tabularx}
	\vspace{-0.5ex} 
	\caption{Quantitative results comparison on different datasets.We mark the \textbf{best} and \underline{second best}.}
	\label{tab:tab_2}
\end{table}

\subsection{Quantitative Evaluation}
The main quantitative results are presented in \cref{tab:tab_2}.
In summary, the proposed method achieves the best overall results on three datasets, followed by AEDNet.
Particularly, DTSNN balances signal retention and noise removal effectively, demonstrating the highest denoising accuracy.
Meanwhile, our model also achieved faster inference speed on average among them. Notably, the runtime from 120 * 120 events indicates that the efficiency of grid-based event representation and processing is significantly higher than the manner based on a single-event level, although the latter better preserves the asynchronous property of the event.
\begin{figure}[t]
	\setlength{\abovecaptionskip}{5pt}
	\setlength{\belowcaptionskip}{-0.4cm}
	\centering
	\includegraphics[width=\linewidth,height=5.8cm]{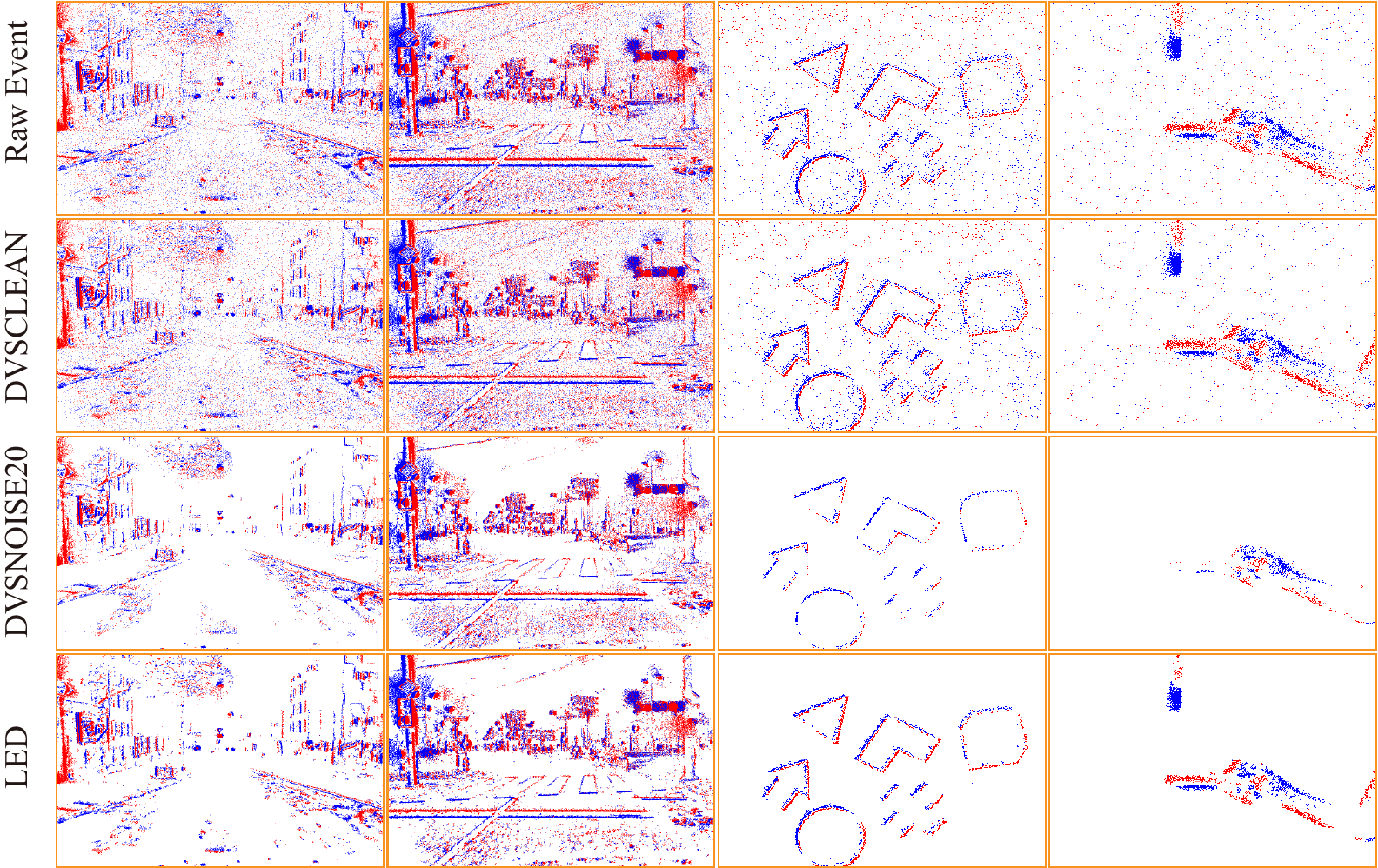}
	\caption{Illustration of the LED diversity. We train DTSNN on different datasets: DVSCLEAN, DVSNOISE20 and LED, and test on real unpaired datasets (from left to right: DSEC→E-MLB). The model trained on LED has achieved better denoising results.}
	\label{fig:fig_9}
\end{figure}
\subsection{Qualitative Evaluation}
\noindent\textbf{Evaluation on LED Dataset}. To further evaluate the denoising performance, we compare with the qualitative results of typical methods on LED.
As shown in \cref{fig:fig_8}, DTSNN outperforms other methods by achieving visually pleasing results, which not only effectively removes noise events in areas such as the sky, ground and building contours, but also preserve the structural features of the scene and objects well. For example, the zoomed region of first row in \cref{fig:fig_8} highlights that our method is the only one successfully preserving the information of small airborne target after denoising. Similarly, our method also stands out in retain  lane marking located in the bottom-left of second row.

\noindent\textbf{Evaluation on Public Datasets}. To evaluate the generalization across different datasets, we also train DTSNN on synthetic event denoising dataset DVSCLEAN \cite{fang2022aednet} and real one DVSNOISE20 \cite{baldwin2020event} respectively, testing on DSEC \cite{Gehrig21ral} and E-MLB \cite{ding2023mlb}, the former is resolution of 640*480 and the latter is resolution of 346*260.
As shown in \cref{fig:fig_9}, the model trained on DVSCLEAN performs poorly due to the huge domain gap between.
DVSNOISE20 excels at removing noise effectively but inevitably loses some details.
The model trained on LED achieves satisfactory results across various scenes in these datasets which simultaneously removes noise without apparent signal loss, strongly supporting the great generalization of the proposed LED dataset.
\begin{figure}[t]
	\setlength{\abovecaptionskip}{5pt}
	\setlength{\belowcaptionskip}{-5pt}
	\centering
	\includegraphics[width=\linewidth]{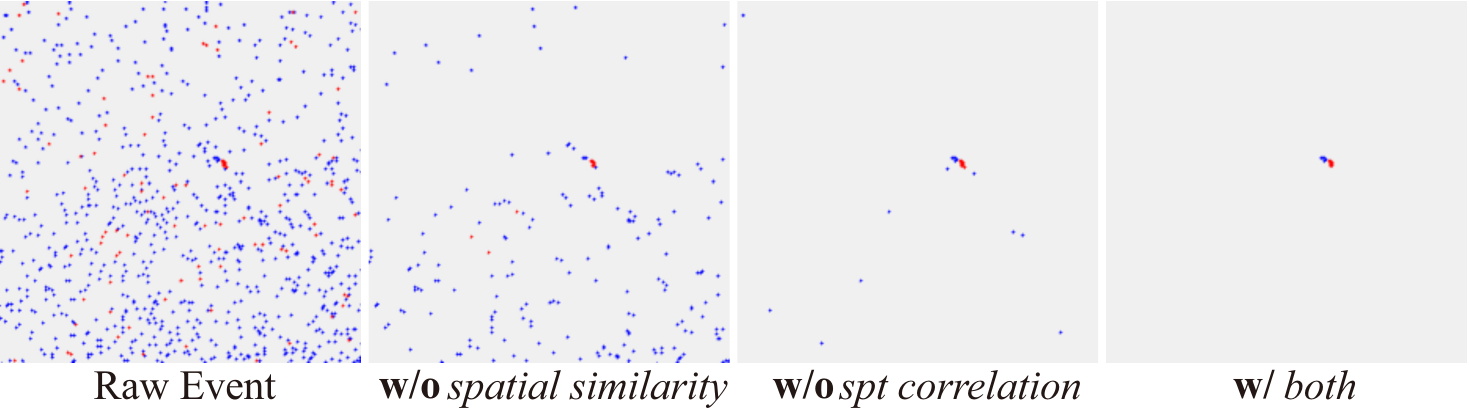}
	\caption{Ablation study of the dual events denoising framework. The first column represents the noisy raw and the remaining three columns represent the flow without spatial similarity, without spatiotemporal correlation constraint and complete DED}
	\label{fig:fig_10}
\end{figure}

\begin{table}[t]
	\setlength{\abovecaptionskip}{5pt}
	\setlength{\belowcaptionskip}{-0.4cm}
	\centering
	\renewcommand{\arraystretch}{0.80}
	\fontsize{8}{12}\selectfont 
	\begin{tabular}{@{}lM{0.35\linewidth}M{0.24\linewidth}@{}}
		\toprule
		Architecture & Denoising Accuracy & Energy Ratio\\
		\midrule
		ANN & 84.5  & $18 \times$ \\
		SNN & 86.0  & $1 \times$ \\
		\addlinespace[-0.5ex] 
		\bottomrule
\end{tabular}
	\vspace{-0.5ex} 
	\caption{Comparison between ANN and SNN.}
	\label{tab:tab_3}
\end{table}

\vspace{-1em}
\subsection{Ablation Study and Discussion}
\noindent\textbf{Effectiveness of DED Framework}. The DED framework aims to fully exploit the property of noise inconsistency /signal consistency in dual events and the self-correlation in a single event stream. As shown in \cref{fig:fig_10}, the small target signal is overwhelmed by noise events, utilizing solely spatiotemporal correlation constraints can only remove partial noise events because some of them also satisfy this relationship. Relying on the spatial similarity of dual event streams effectively disassembles the mutual relationship in the dense noise event group. Therefore, combining both procedures can achieve a more refined denoising result.

\noindent\textbf{Effectiveness of Spiking Neuron}. Intuitively, the inherent spatiotemporal sparsity of event streams seamlessly aligns with SNN because of the temporal information ability of spiking neurons.
To compare the performance differences between SNN and ANN architectures in the event denoising tasks, the evaluation was conducted on the same structure.
As shown in \cref{tab:tab_3}, the SNN based on LIF neuron achieves better denoising accuracy while maintaining $18 \times$ lower power consumption advantage compared to its ANN version (the spiking neurons replaced with ReLU).

\begin{figure}[t]
	\setlength{\abovecaptionskip}{5pt}
	\setlength{\belowcaptionskip}{-3pt}
	\centering
	\includegraphics[width=\linewidth]{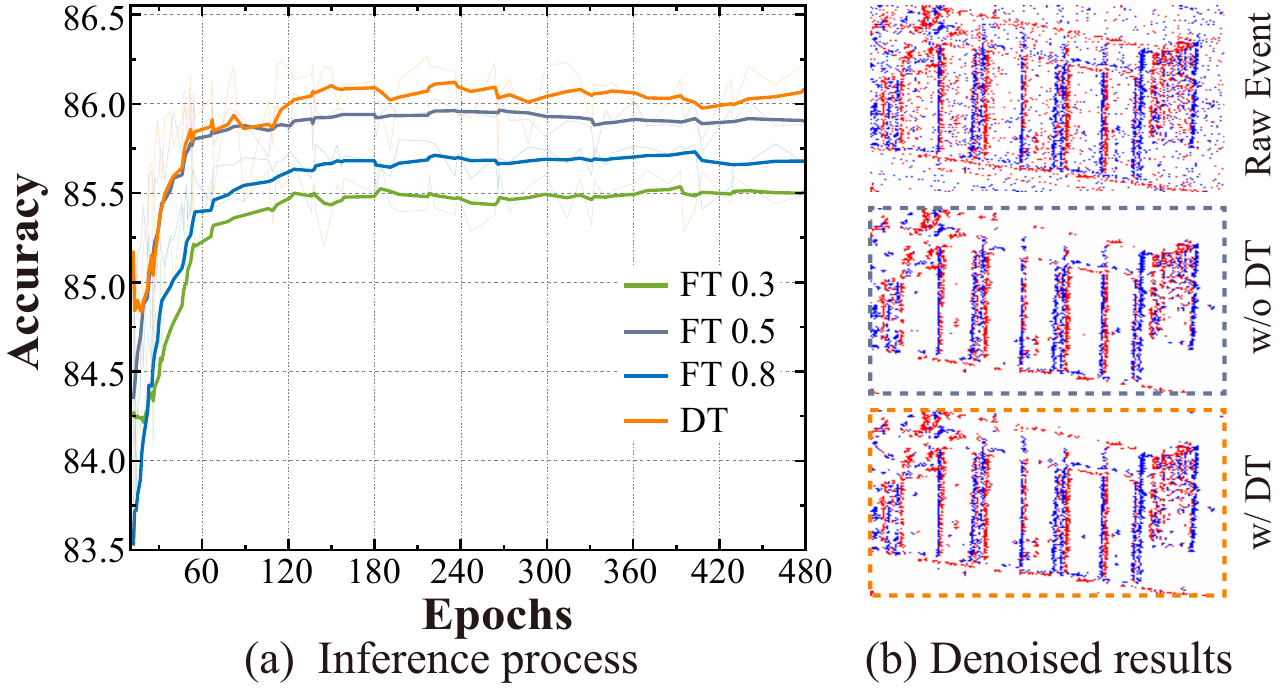}
	\caption{Illustration of the DT model Effectiveness. (a) Inference accuracy of SNN with various fixed thresholds and with DT. (b) Denoised results without DT and with DT, respectively.}
	\label{fig:fig_11}
\end{figure}

\noindent\textbf{Effectiveness of DT Module}. To evaluate the effects of the proposed learnable threshold mechanism of DTSNN, we train the network with various fixed threshold neurons and dynamic threshold neurons, respectively.
As can be observed from \cref{tab:tab_4} and \cref{fig:fig_11} (a), the denoising accuracy improves after introducing the DT module due to the network has the tendency to increasing spiking probability for signal pixels according to threshold prediction, and the \cref{fig:fig_11} (b) illustrates DT module could assist in locating the approximate signal or noise region to retain more signal events from structural area compared to the FT model.

\begin{table}[t]
	\setlength{\abovecaptionskip}{5pt}
	\setlength{\belowcaptionskip}{-0.3cm}
	\centering
	\renewcommand{\arraystretch}{0.80}
	\fontsize{8}{12}\selectfont 
	\begin{tabular}{@{}lM{0.16\linewidth}M{0.15\linewidth}M{0.15\linewidth}M{0.15\linewidth}@{}}
		\toprule
		Type & FTSNN (0.3 th) & FTSNN (0.5 th)  &  FTSNN  (0.8 th)  &  DTSNN \\[0.15cm]
		\midrule
		Accuracy & 85.5 & 86.0 & 85.7 & 86.2 \\
		\addlinespace[-0.5ex] 
\bottomrule
\end{tabular}
\vspace{-0.5ex} 
	\caption{Inference accuracy with proposed methods implemented.}
	\label{tab:tab_4}
\end{table}

\noindent\textbf{Discussion}. Essentially, true events comes from intensity changes. When artificial light source variations exist, the light source itself, its coverage area and reflection from ground texture would result in relevant events. These signal events exhibit noise-like features with discontinuous structure, while the proposed DED framework considers them as signal event that should be preserved. Thus, with the supervision of such signal labels, the proposed model also tend to retain such events arise from lamp as shown in left two columns of \cref{fig:fig_9} despite their visually unappealing effect. 

\noindent\textbf{Limitation}. The DED framework may fail in extreme cases where similar part are dominated by noise events. However, such globally anomalous high-frequency noise is rare in practice. Moreover, the proposed model struggles to capture those insignificant signals attaching to structural periphery, limiting signal retain performance in metrics.
\section{Conclusion}
In this paper, we construct a new large-scale high-quality paired real event denoising dataset LED, which provides diverse noise levels and scenes which aims to cover various internal and external conditions.
Based on the consistency difference of noise and signal events, the proposed DED framework could effectively generate high-quality GT and detailed analysis confirms our better results than others.
In addition, we propose a novel baseline DTSNN for event denoising featuring with dynamic threshold mechanism of LIF neuron.
The experimental results demonstrate the superiority of the proposed dataset and denoising method.

\noindent\textbf{Acknowledgments}. The computation is completed in the HPC Platform of Huazhong University of Science and Technology. This work was supported by the National Natural Science Foundation of China under Grant 62371203.

\clearpage
{
    \small
    \bibliographystyle{ieeenat_fullname}
    \bibliography{main}

\begin{thebibliography}{52}
\providecommand{\natexlab}[1]{#1}
\providecommand{\url}[1]{\texttt{#1}}
\expandafter\ifx\csname urlstyle\endcsname\relax
  \providecommand{\doi}[1]{doi: #1}\else
  \providecommand{\doi}{doi: \begingroup \urlstyle{rm}\Url}\fi

\bibitem[Alkendi et~al.(2022)Alkendi, Azzam, Ayyad, Javed, Seneviratne, and
  Zweiri]{alkendi2022}
Yusra Alkendi, Rana Azzam, Abdulla Ayyad, Sajid Javed, Lakmal Seneviratne, and
  Yahya Zweiri.
\newblock Neuromorphic camera denoising using graph neural network-driven
  transformers.
\newblock \emph{IEEE Transactions on Neural Networks and Learning Systems},
  2022.

\bibitem[Baldwin et~al.(2020)Baldwin, Almatrafi, Asari, and
  Hirakawa]{baldwin2020event}
R Baldwin, Mohammed Almatrafi, Vijayan Asari, and Keigo Hirakawa.
\newblock Event probability mask (epm) and event denoising convolutional neural
  network (edncnn) for neuromorphic cameras.
\newblock In \emph{Proceedings of the IEEE/CVF Conference on Computer Vision
  and Pattern Recognition}, pages 1701--1710, 2020.

\bibitem[Benosman et~al.(2013)Benosman, Clercq, Lagorce, Ieng, and
  Bartolozzi]{benosman2013event}
Ryad Benosman, Charles Clercq, Xavier Lagorce, Sio-Hoi Ieng, and Chiara
  Bartolozzi.
\newblock Event-based visual flow.
\newblock \emph{IEEE transactions on neural networks and learning systems},
  25\penalty0 (2):\penalty0 407--417, 2013.

\bibitem[Cabriel et~al.(2023)Cabriel, Monfort, Specht, and
  Izeddin]{cabriel2023event}
Cl{\'e}ment Cabriel, Tual Monfort, Christian~G Specht, and Ignacio Izeddin.
\newblock Event-based vision sensor for fast and dense single-molecule
  localization microscopy.
\newblock \emph{Nature Photonics}, pages 1--9, 2023.

\bibitem[Chen et~al.(2020)Chen, Suter, Wu, and Wang]{chen2020end}
Haosheng Chen, David Suter, Qiangqiang Wu, and Hanzi Wang.
\newblock End-to-end learning of object motion estimation from retinal events
  for event-based object tracking.
\newblock In \emph{Proceedings of the AAAI Conference on Artificial
  Intelligence}, pages 10534--10541, 2020.

\bibitem[Chen et~al.(2022)Chen, Wang, Cao, Wu, and
  Zha]{chen2022progressivemotionseg}
Jinze Chen, Yang Wang, Yang Cao, Feng Wu, and Zheng-Jun Zha.
\newblock Progressivemotionseg: Mutually reinforced framework for event-based
  motion segmentation.
\newblock In \emph{Proceedings of the AAAI Conference on Artificial
  Intelligence}, pages 303--311, 2022.

\bibitem[Ding et~al.(2022)Ding, Dong, Heide, Ding, Zhou, Yin, and
  Yang]{ding2022biologically}
Jianchuan Ding, Bo Dong, Felix Heide, Yufei Ding, Yunduo Zhou, Baocai Yin, and
  Xin Yang.
\newblock Biologically inspired dynamic thresholds for spiking neural networks.
\newblock \emph{Advances in Neural Information Processing Systems},
  35:\penalty0 6090--6103, 2022.

\bibitem[Ding et~al.(2023)Ding, Chen, Wang, Kang, Song, Cheng, and
  Cao]{ding2023mlb}
Saizhe Ding, Jinze Chen, Yang Wang, Yu Kang, Weiguo Song, Jie Cheng, and Yang
  Cao.
\newblock E-mlb: Multilevel benchmark for event-based camera denoising.
\newblock \emph{IEEE Transactions on Multimedia}, 2023.

\bibitem[Duan et~al.(2021{\natexlab{a}})Duan, Wang, Shi, Cossairt, Huang, and
  Katsaggelos]{duan2021guided}
Peiqi Duan, Zihao~W Wang, Boxin Shi, Oliver Cossairt, Tiejun Huang, and
  Aggelos~K Katsaggelos.
\newblock Guided event filtering: Synergy between intensity images and
  neuromorphic events for high performance imaging.
\newblock \emph{IEEE Transactions on Pattern Analysis and Machine
  Intelligence}, 44\penalty0 (11):\penalty0 8261--8275, 2021{\natexlab{a}}.

\bibitem[Duan et~al.(2021{\natexlab{b}})Duan, Wang, Zhou, Ma, and
  Shi]{duan2021eventzoom}
Peiqi Duan, Zihao~W Wang, Xinyu Zhou, Yi Ma, and Boxin Shi.
\newblock Eventzoom: Learning to denoise and super resolve neuromorphic events.
\newblock In \emph{Proceedings of the IEEE/CVF Conference on Computer Vision
  and Pattern Recognition}, pages 12824--12833, 2021{\natexlab{b}}.

\bibitem[Falanga et~al.(2020)Falanga, Kleber, and
  Scaramuzza]{falanga2020dynamic}
Davide Falanga, Kevin Kleber, and Davide Scaramuzza.
\newblock Dynamic obstacle avoidance for quadrotors with event cameras.
\newblock \emph{Science Robotics}, 5\penalty0 (40):\penalty0 eaaz9712, 2020.

\bibitem[Fang et~al.(2022)Fang, Wu, Li, Hou, Dong, and Shi]{fang2022aednet}
Huachen Fang, Jinjian Wu, Leida Li, Junhui Hou, Weisheng Dong, and Guangming
  Shi.
\newblock Aednet: Asynchronous event denoising with spatial-temporal
  correlation among irregular data.
\newblock In \emph{Proceedings of the 30th ACM International Conference on
  Multimedia}, pages 1427--1435, 2022.

\bibitem[Fang et~al.(2021)Fang, Yu, Chen, Masquelier, Huang, and
  Tian]{fang2021incorporating}
Wei Fang, Zhaofei Yu, Yanqi Chen, Timoth{\'e}e Masquelier, Tiejun Huang, and
  Yonghong Tian.
\newblock Incorporating learnable membrane time constant to enhance learning of
  spiking neural networks.
\newblock In \emph{Proceedings of the IEEE/CVF international conference on
  computer vision}, pages 2661--2671, 2021.

\bibitem[Fang et~al.(2023)Fang, Chen, Ding, Yu, Masquelier, Chen, Huang, Zhou,
  Li, and Tian]{fang2023spikingjelly}
Wei Fang, Yanqi Chen, Jianhao Ding, Zhaofei Yu, Timoth{\'e}e Masquelier, Ding
  Chen, Liwei Huang, Huihui Zhou, Guoqi Li, and Yonghong Tian.
\newblock Spikingjelly: An open-source machine learning infrastructure platform
  for spike-based intelligence.
\newblock \emph{Science Advances}, 9\penalty0 (40):\penalty0 eadi1480, 2023.

\bibitem[Feng et~al.(2020)Feng, Lv, Liu, Zhang, Xiao, and Han]{feng2020event}
Yang Feng, Hengyi Lv, Hailong Liu, Yisa Zhang, Yuyao Xiao, and Chengshan Han.
\newblock Event density based denoising method for dynamic vision sensor.
\newblock \emph{Applied Sciences}, 10\penalty0 (6):\penalty0 2024, 2020.

\bibitem[Gallego et~al.(2018)Gallego, Rebecq, and
  Scaramuzza]{gallego2018unifying}
Guillermo Gallego, Henri Rebecq, and Davide Scaramuzza.
\newblock A unifying contrast maximization framework for event cameras, with
  applications to motion, depth, and optical flow estimation.
\newblock In \emph{Proceedings of the IEEE conference on computer vision and
  pattern recognition}, pages 3867--3876, 2018.

\bibitem[Gehrig et~al.(2020)Gehrig, Gehrig, Hidalgo-Carri{\'o}, and
  Scaramuzza]{gehrig2020video}
Daniel Gehrig, Mathias Gehrig, Javier Hidalgo-Carri{\'o}, and Davide
  Scaramuzza.
\newblock Video to events: Recycling video datasets for event cameras.
\newblock In \emph{Proceedings of the IEEE/CVF Conference on Computer Vision
  and Pattern Recognition}, pages 3586--3595, 2020.

\bibitem[Gehrig et~al.(2021)Gehrig, Aarents, Gehrig, and
  Scaramuzza]{Gehrig21ral}
Mathias Gehrig, Willem Aarents, Daniel Gehrig, and Davide Scaramuzza.
\newblock Dsec: A stereo event camera dataset for driving scenarios.
\newblock \emph{IEEE Robotics and Automation Letters}, 2021.

\bibitem[Gerstner et~al.(2014)Gerstner, Kistler, Naud, and
  Paninski]{gerstner2014neuronal}
Wulfram Gerstner, Werner~M Kistler, Richard Naud, and Liam Paninski.
\newblock \emph{Neuronal dynamics: From single neurons to networks and models
  of cognition}.
\newblock Cambridge University Press, 2014.

\bibitem[Guo and Delbruck(2022)]{guo2022low}
Shasha Guo and Tobi Delbruck.
\newblock Low cost and latency event camera background activity denoising.
\newblock \emph{IEEE Transactions on Pattern Analysis and Machine
  Intelligence}, 45\penalty0 (1):\penalty0 785--795, 2022.

\bibitem[Hasinoff et~al.(2016)Hasinoff, Sharlet, Geiss, Adams, Barron, Kainz,
  Chen, and Levoy]{hasinoff2016burst}
Samuel~W Hasinoff, Dillon Sharlet, Ryan Geiss, Andrew Adams, Jonathan~T Barron,
  Florian Kainz, Jiawen Chen, and Marc Levoy.
\newblock Burst photography for high dynamic range and low-light imaging on
  mobile cameras.
\newblock \emph{ACM Transactions on Graphics (ToG)}, 35\penalty0 (6):\penalty0
  1--12, 2016.

\bibitem[Hu et~al.(2021)Hu, Liu, and Delbruck]{hu2021v2e}
Yuhuang Hu, Shih-Chii Liu, and Tobi Delbruck.
\newblock v2e: From video frames to realistic dvs events.
\newblock In \emph{Proceedings of the IEEE/CVF Conference on Computer Vision
  and Pattern Recognition}, pages 1312--1321, 2021.

\bibitem[Huang et~al.(2021)Huang, Li, Jia, Lu, and
  Liu]{huang2021neighbor2neighbor}
Tao Huang, Songjiang Li, Xu Jia, Huchuan Lu, and Jianzhuang Liu.
\newblock Neighbor2neighbor: Self-supervised denoising from single noisy
  images.
\newblock In \emph{Proceedings of the IEEE/CVF conference on computer vision
  and pattern recognition}, pages 14781--14790, 2021.

\bibitem[Khodamoradi and Kastner(2018)]{Knoise2018}
Alireza Khodamoradi and Ryan Kastner.
\newblock $o(n)$o(n)-space spatiotemporal filter for reducing noise in
  neuromorphic vision sensors.
\newblock \emph{IEEE Transactions on Emerging Topics in Computing}, 9\penalty0
  (1):\penalty0 15--23, 2018.

\bibitem[Krull et~al.(2019)Krull, Buchholz, and Jug]{krull2019noise2void}
Alexander Krull, Tim-Oliver Buchholz, and Florian Jug.
\newblock Noise2void-learning denoising from single noisy images.
\newblock In \emph{Proceedings of the IEEE/CVF conference on computer vision
  and pattern recognition}, pages 2129--2137, 2019.

\bibitem[Lagorce et~al.(2017)Lagorce, Orchard, Galluppi, Shi, and
  Benosman]{Lagorce2017}
Xavier Lagorce, Garrick Orchard, Francesco Galluppi, Bertram~E. Shi, and
  Ryad~B. Benosman.
\newblock Hots: A hierarchy of event-based time-surfaces for pattern
  recognition.
\newblock \emph{IEEE Transactions on Pattern Analysis and Machine
  Intelligence}, 39\penalty0 (7):\penalty0 1346--1359, 2017.

\bibitem[Lehtinen et~al.(2018)Lehtinen, Munkberg, Hasselgren, Laine, Karras,
  Aittala, and Aila]{lehtinen2018noise2noise}
Jaakko Lehtinen, Jacob Munkberg, Jon Hasselgren, Samuli Laine, Tero Karras,
  Miika Aittala, and Timo Aila.
\newblock Noise2noise: Learning image restoration without clean data.
\newblock \emph{arXiv preprint arXiv:1803.04189}, 2018.

\bibitem[Liba et~al.(2019)Liba, Murthy, Tsai, Brooks, Xue, Karnad, He, Barron,
  Sharlet, Geiss, et~al.]{liba2019handheld}
Orly Liba, Kiran Murthy, Yun-Ta Tsai, Tim Brooks, Tianfan Xue, Nikhil Karnad,
  Qiurui He, Jonathan~T Barron, Dillon Sharlet, Ryan Geiss, et~al.
\newblock Handheld mobile photography in very low light.
\newblock \emph{ACM Trans. Graph.}, 38\penalty0 (6):\penalty0 164--1, 2019.

\bibitem[Lichtsteiner et~al.(2008)Lichtsteiner, Posch, and
  Delbruck]{lichtsteiner2008}
Patrick Lichtsteiner, Christoph Posch, and Tobi Delbruck.
\newblock A 128$\times$128 120 db 15 $\mu$s latency asynchronous temporal
  contrast vision sensor.
\newblock \emph{IEEE journal of solid-state circuits}, 43\penalty0
  (2):\penalty0 566--576, 2008.

\bibitem[Liu et~al.(2015)Liu, Brandli, Li, Liu, and Delbruck]{liu2015design}
Hongjie Liu, Christian Brandli, Chenghan Li, Shih-Chii Liu, and Tobi Delbruck.
\newblock Design of a spatiotemporal correlation filter for event-based
  sensors.
\newblock In \emph{2015 IEEE International Symposium on Circuits and Systems
  (ISCAS)}, pages 722--725. IEEE, 2015.

\bibitem[Liu et~al.(2014)Liu, Yuan, Tang, Uyttendaele, and Sun]{liu2014fast}
Ziwei Liu, Lu Yuan, Xiaoou Tang, Matt Uyttendaele, and Jian Sun.
\newblock Fast burst images denoising.
\newblock \emph{ACM Transactions on Graphics (TOG)}, 33\penalty0 (6):\penalty0
  1--9, 2014.

\bibitem[Maleky et~al.(2022)Maleky, Kousha, Brown, and
  Brubaker]{maleky2022noise2noiseflow}
Ali Maleky, Shayan Kousha, Michael~S Brown, and Marcus~A Brubaker.
\newblock Noise2noiseflow: realistic camera noise modeling without clean
  images.
\newblock In \emph{Proceedings of the IEEE/CVF Conference on Computer Vision
  and Pattern Recognition}, pages 17632--17641, 2022.

\bibitem[Mangalwedhekar et~al.(2023)Mangalwedhekar, Singh, Thakur,
  Seelamantula, Jose, and Nair]{mangalwedhekar2023achieving}
Rohit Mangalwedhekar, Nivedita Singh, Chetan~Singh Thakur, Chandra~Sekhar
  Seelamantula, Mini Jose, and Deepak Nair.
\newblock Achieving nanoscale precision using neuromorphic localization
  microscopy.
\newblock \emph{Nature Nanotechnology}, pages 1--10, 2023.

\bibitem[Mildenhall et~al.(2018)Mildenhall, Barron, Chen, Sharlet, Ng, and
  Carroll]{mildenhall2018burst}
Ben Mildenhall, Jonathan~T Barron, Jiawen Chen, Dillon Sharlet, Ren Ng, and
  Robert Carroll.
\newblock Burst denoising with kernel prediction networks.
\newblock In \emph{Proceedings of the IEEE conference on computer vision and
  pattern recognition}, pages 2502--2510, 2018.

\bibitem[Mittal et~al.(2012)Mittal, Soundararajan, and Bovik]{mittal2012making}
Anish Mittal, Rajiv Soundararajan, and Alan~C Bovik.
\newblock Making a “completely blind” image quality analyzer.
\newblock \emph{IEEE Signal processing letters}, 20\penalty0 (3):\penalty0
  209--212, 2012.

\bibitem[Pan et~al.(2020)Pan, Hartley, Scheerlinck, Liu, Yu, and
  Dai]{pan2020high}
Liyuan Pan, Richard Hartley, Cedric Scheerlinck, Miaomiao Liu, Xin Yu, and
  Yuchao Dai.
\newblock High frame rate video reconstruction based on an event camera.
\newblock \emph{IEEE Transactions on Pattern Analysis and Machine
  Intelligence}, 44\penalty0 (5):\penalty0 2519--2533, 2020.

\bibitem[Perot et~al.(2020)Perot, De~Tournemire, Nitti, Masci, and
  Sironi]{perot2020learning}
Etienne Perot, Pierre De~Tournemire, Davide Nitti, Jonathan Masci, and Amos
  Sironi.
\newblock Learning to detect objects with a 1 megapixel event camera.
\newblock \emph{Advances in Neural Information Processing Systems},
  33:\penalty0 16639--16652, 2020.

\bibitem[Rebecq et~al.(2018)Rebecq, Gehrig, and Scaramuzza]{rebecq2018esim}
Henri Rebecq, Daniel Gehrig, and Davide Scaramuzza.
\newblock Esim: an open event camera simulator.
\newblock In \emph{Conference on robot learning}, pages 969--982, 2018.

\bibitem[Rebecq et~al.(2019)Rebecq, Ranftl, Koltun, and
  Scaramuzza]{rebecq2019high}
Henri Rebecq, Ren{\'e} Ranftl, Vladlen Koltun, and Davide Scaramuzza.
\newblock High speed and high dynamic range video with an event camera.
\newblock \emph{IEEE transactions on pattern analysis and machine
  intelligence}, 43\penalty0 (6):\penalty0 1964--1980, 2019.

\bibitem[Stoffregen and Kleeman(2019)]{stoffregen2019reward}
Timo Stoffregen and Lindsay Kleeman.
\newblock Event cameras, contrast maximization and reward functions: An
  analysis.
\newblock In \emph{Proceedings of the IEEE/CVF Conference on Computer Vision
  and Pattern Recognition}, pages 12300--12308, 2019.

\bibitem[Stoffregen et~al.(2019)Stoffregen, Gallego, Drummond, Kleeman, and
  Scaramuzza]{stoffregen2019event}
Timo Stoffregen, Guillermo Gallego, Tom Drummond, Lindsay Kleeman, and Davide
  Scaramuzza.
\newblock Event-based motion segmentation by motion compensation.
\newblock In \emph{Proceedings of the IEEE/CVF International Conference on
  Computer Vision}, pages 7244--7253, 2019.

\bibitem[Systems()]{cadence}
Cadence~Design Systems.
\newblock Software downloads-cadence design systems.
\newblock
  \url{https://www.cadence.com/en_US/home/support/software-downloads.html}.

\bibitem[Wang et~al.(2022)Wang, Cheng, and Lim]{wang2022ltmd}
Siqi Wang, Tee~Hiang Cheng, and Meng-Hiot Lim.
\newblock Ltmd: Learning improvement of spiking neural networks with learnable
  thresholding neurons and moderate dropout.
\newblock \emph{Advances in Neural Information Processing Systems},
  35:\penalty0 28350--28362, 2022.

\bibitem[Wang et~al.(2019)Wang, Du, Shen, Wu, Zhao, Sun, and Wen]{wang2019ev}
Yanxiang Wang, Bowen Du, Yiran Shen, Kai Wu, Guangrong Zhao, Jianguo Sun, and
  Hongkai Wen.
\newblock Ev-gait: Event-based robust gait recognition using dynamic vision
  sensors.
\newblock In \emph{Proceedings of the IEEE/CVF Conference on Computer Vision
  and Pattern Recognition}, pages 6358--6367, 2019.

\bibitem[Wang et~al.(2020)Wang, Duan, Cossairt, Katsaggelos, Huang, and
  Shi]{wang2020joint}
Zihao~W Wang, Peiqi Duan, Oliver Cossairt, Aggelos Katsaggelos, Tiejun Huang,
  and Boxin Shi.
\newblock Joint filtering of intensity images and neuromorphic events for
  high-resolution noise-robust imaging.
\newblock In \emph{Proceedings of the IEEE/CVF Conference on Computer Vision
  and Pattern Recognition}, pages 1609--1619, 2020.

\bibitem[Wei et~al.(2020)Wei, Fu, Yang, and Huang]{wei2020physics}
Kaixuan Wei, Ying Fu, Jiaolong Yang, and Hua Huang.
\newblock A physics-based noise formation model for extreme low-light raw
  denoising.
\newblock In \emph{Proceedings of the IEEE/CVF Conference on Computer Vision
  and Pattern Recognition}, pages 2758--2767, 2020.

\bibitem[Wu et~al.(2020)Wu, Ma, Li, Dong, and Shi]{wu2020probabilistic}
Jinjian Wu, Chuanwei Ma, Leida Li, Weisheng Dong, and Guangming Shi.
\newblock Probabilistic undirected graph based denoising method for dynamic
  vision sensor.
\newblock \emph{IEEE Transactions on Multimedia}, 23:\penalty0 1148--1159,
  2020.

\bibitem[Yao et~al.(2021)Yao, Gao, Zhao, Wang, Lin, Yang, and
  Li]{yao2021temporal}
Man Yao, Huanhuan Gao, Guangshe Zhao, Dingheng Wang, Yihan Lin, Zhaoxu Yang,
  and Guoqi Li.
\newblock Temporal-wise attention spiking neural networks for event streams
  classification.
\newblock In \emph{Proceedings of the IEEE/CVF International Conference on
  Computer Vision}, pages 10221--10230, 2021.

\bibitem[Zhang et~al.(2022)Zhang, Dong, Zhang, Ding, Heide, Yin, and
  Yang]{zhang2022spiking}
Jiqing Zhang, Bo Dong, Haiwei Zhang, Jianchuan Ding, Felix Heide, Baocai Yin,
  and Xin Yang.
\newblock Spiking transformers for event-based single object tracking.
\newblock In \emph{Proceedings of the IEEE/CVF conference on Computer Vision
  and Pattern Recognition}, pages 8801--8810, 2022.

\bibitem[Zhang et~al.(2023)Zhang, Ge, Song, and L]{zhang2023neuromorphic}
Pei Zhang, Zhou Ge, Li Song, and Edmund~Y L.
\newblock Neuromorphic imaging with density-based spatiotemporal denoising.
\newblock \emph{IEEE Transactions on Computational Imaging}, 2023.

\bibitem[Zhang et~al.(2021)Zhang, Liao, Yu, Yang, and Xia]{zhang2021event}
Xiang Zhang, Wei Liao, Lei Yu, Wen Yang, and Gui-Song Xia.
\newblock Event-based synthetic aperture imaging with a hybrid network.
\newblock In \emph{Proceedings of the IEEE/CVF Conference on Computer Vision
  and Pattern Recognition}, pages 14235--14244, 2021.

\bibitem[Zhu et~al.(2022)Zhu, Wang, Chang, Li, Huang, and Tian]{zhu2022event}
Lin Zhu, Xiao Wang, Yi Chang, Jianing Li, Tiejun Huang, and Yonghong Tian.
\newblock Event-based video reconstruction via potential-assisted spiking
  neural network.
\newblock In \emph{Proceedings of the IEEE/CVF Conference on Computer Vision
  and Pattern Recognition}, pages 3594--3604, 2022.

\end{thebibliography}
}

\end{document}